\pdfoutput=1
\documentclass{llncs}
\usepackage{llncsdoc}
\usepackage{amsmath}
\usepackage{graphicx}
\usepackage{amsfonts}
\usepackage{amsmath}
\usepackage{xcolor}
\usepackage{float}
\usepackage{subcaption}

\newcommand{\eg}{e.\,g.}
\newcommand{\ie}{i.\,e.}
\newcommand{\etal}{\emph{et al.}}

\begin{document}
\title{Reconstructing Sinus Anatomy from Endoscopic Video -- Towards a Radiation-free Approach for Quantitative Longitudinal Assessment}
\author{Xingtong Liu \inst{1} \and Maia Stiber \inst{1} \and Jindan Huang \inst{1} \and Masaru Ishii \inst{2} \and Gregory D. Hager \inst{1} \and Russell H. Taylor \inst{1} \and Mathias Unberath \inst{1}} 

\institute{The Johns Hopkins University, Baltimore, USA,\\ \email{\{xingtongliu,unberath\}@jhu.edu}
\and
Johns Hopkins Medical Institutions, Baltimore, USA}

\maketitle

\begin{abstract}
Reconstructing accurate 3D surface models of sinus anatomy directly from an endoscopic video is a promising avenue for cross-sectional and longitudinal analysis to better understand the relationship between sinus anatomy and surgical outcomes. We present a patient-specific, learning-based method for 3D reconstruction of sinus surface anatomy directly and only from endoscopic videos. We demonstrate the effectiveness and accuracy of our method on \emph{in} and \emph{ex vivo} data where we compare to sparse reconstructions from Structure from Motion, dense reconstruction from COLMAP, and ground truth anatomy from CT. Our textured reconstructions are watertight and enable measurement of clinically relevant parameters in good agreement with CT. The source code is available at \url{https://github.com/lppllppl920/DenseReconstruction-Pytorch}.
\end{abstract}

\section{Introduction}
The prospect of reconstructing accurate 3D surface models of sinus anatomy directly from endoscopic videos is exciting in multiple regards. Many diseases are defined by aberrations in human geometry, such as laryngotracheal stenosis, obstructive sleep apnea, and nasal obstruction in the head and neck region. In these diseases, patients suffer significantly due to the narrowing of the airway. While billions of dollars are spent to manage these patients, the outcomes are not exclusively satisfactory. An example: The two most common surgeries for nasal obstruction, septoplasty and turbinate reduction, are generally reported to \emph{on average} significantly improve disease-specific quality of life~\cite{bezerra2012quality}, but evidence suggests that these improvements are short term in more than $40\,$\% of cases~\cite{hytonen2006we,hytonen2012does}. Some hypotheses attribute the low success rate to anatomical geometry but there are no objective measures to support these claims. The ability to analyze longitudinal geometric data from a large population will potentially help to better understand the relationship between sinus anatomy and surgical outcomes. In current practice, CT is the gold standard for obtaining accurate 3D information about patient anatomy. However, due to its high cost and use of ionizing radiation, CT scanning is not suitable for longitudinal monitoring of patient anatomy. Endoscopy is routinely performed in outpatient and clinic settings to qualitatively assess treatment effect, and thus, constitutes an ideal modality to collect longitudinal data. In order to use endoscopic video data to analyze and model sinus anatomy in 3D, methods for 3D surface reconstruction that operate solely on endoscopic video are required. The resulting 3D reconstructions must agree with CT and allow for geometric measurement of clinically relevant parameters, \eg~aperture and volume.

\textbf{Contributions.}\quad To address these challenges, we propose a patient-specific learning-based method for 3D sinus surface reconstruction from endoscopic videos. Our textured reconstructions are watertight and enable measurement of clinically relevant parameters in good agreement with CT. We extensively demonstrate the effectiveness and accuracy of our method on \emph{in} and \emph{ex vivo} data where we compare to sparse reconstructions from Structure from Motion (SfM), dense reconstruction from COLMAP~\cite{Schonberger2016COLMAP}, and ground truth anatomy from CT. 

\textbf{Related Work.}\quad Many methods to estimate surface reconstruction from endoscopic videos have been proposed. SfM-based methods aim for texture smoothness~\cite{Phan2019DenseOpticalFlow,Widya2019WholeStomach,Qiu2018EndoscopeOral} and provide a sparse or dense reconstructed point cloud that is then processed by a surface reconstruction method, such as Poisson reconstruction~\cite{kazhdan2006poisson}. Unfortunately, there are no guarantees that this approach will result in reasonable surfaces specifically when applied to anatomically complex structures, such as the nasal cavity in Fig.~\ref{fig:reconstruction_method_comparison}. Shape-from-Shading methods are often combined with fusion techniques, such as \cite{Turan2018Sparse-Then-Dense,Tokgozoglu2012ColorHybrid,Karargyris2011ThreeDigestive}, and often require careful photometric calibration to ensure accuracy. Reconstruction with tissue deformation are handled in~\cite{Zhao2016Endoscopogram,lamarca2019defslam,song2018mis}. In intra-operative scenarios, SLAM-based methods~\cite{lamarca2019defslam,song2018mis,ma2019real} are preferable as they optimize for near real-time execution. Learning-based methods~\cite{ma2019real,chen2019slam} take advantage of deep learning advancements in depth and pose estimation to improve model quality. 

\section{Methods}
\begin{figure*}[t]
	\centering
	\includegraphics[width=0.9\textwidth]{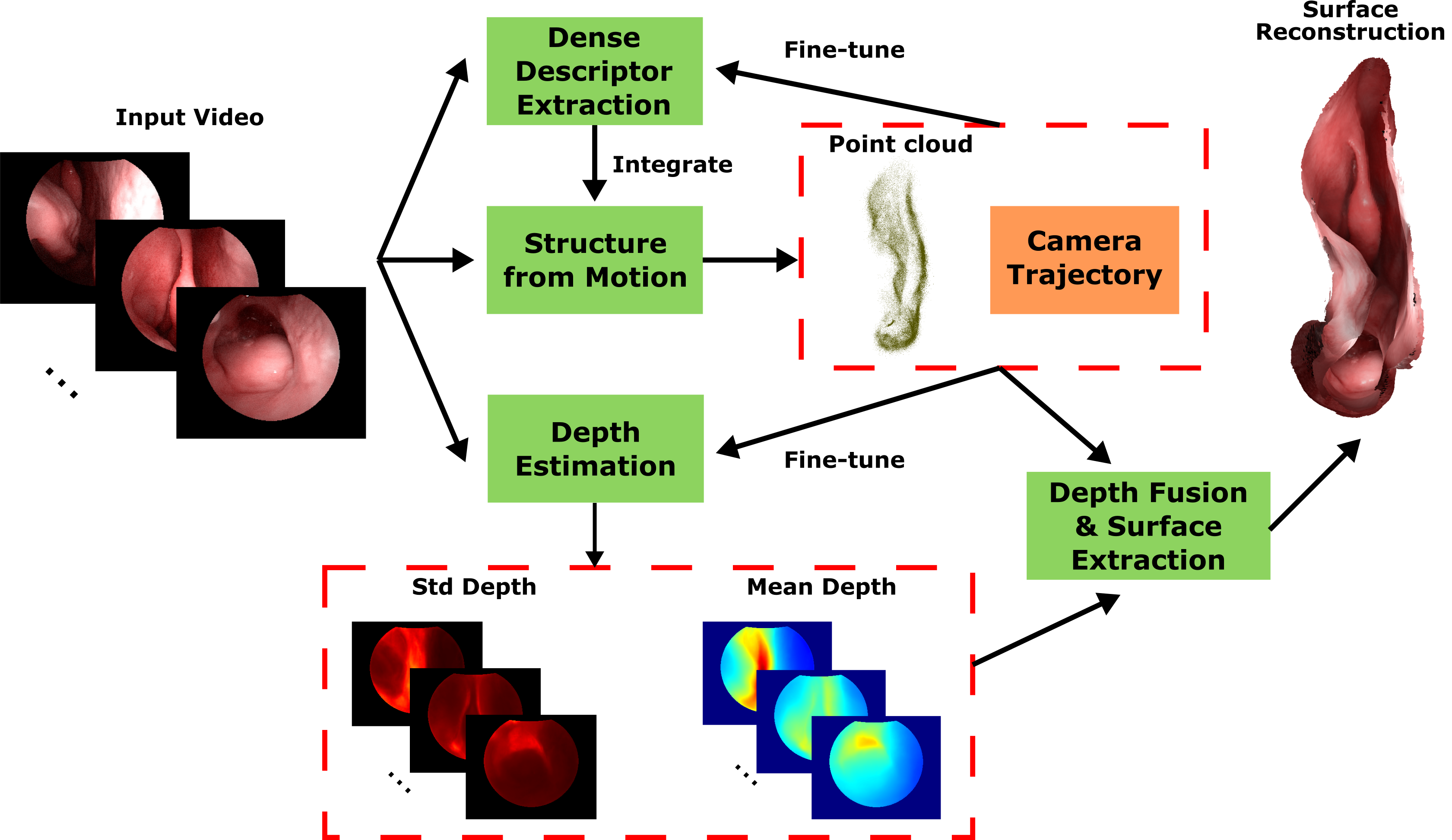}
	\caption{\textbf{Overall pipeline.} Note that part of the surface reconstructions is removed in the figure to display internal structures. 
	}
	\label{fig:overall_pipeline}
\end{figure*}

\textbf{Overall Pipeline.}\quad The goal of our proposed pipeline is to automatically reconstruct a watertight textured sinus surface from an unlabeled endoscopic video. The pipeline, shown in Fig.~\ref{fig:overall_pipeline}, has three main components: 1) SfM based on dense point correspondences produced by a learning-based descriptor; 2) depth estimation; and 3) volumetric depth fusion with surface extraction. SfM identifies corresponding points across the video sequence and uses these correspondences to calculate both, a sparse 3D reconstruction of these points and the camera trajectory generating the video. By replacing local with learning-based descriptors during the point correspondence stage of SfM, we are able to improve the density of sparse reconstruction and the completeness of the estimated camera trajectory. We refer to this process as \emph{Dense Descriptor Extraction}. Reliable and complete reconstructions directly from SfM are important, because these results are subsequently used for two purposes: 1) they provide self-supervisory signals for fine-tuning two learning-based modules, \ie~\emph{Dense Descriptor Extraction} and \emph{Depth Estimation}; 2) they are used in the \emph{Depth Fusion \& Surface Extraction} module to guide the fusion procedure. \emph{Depth Estimation} provides dense depth measurements for all pixels in every video frame that are then aggregated over the whole video sequence using \emph{Depth Fusion \& Surface Extraction}. 

\textbf{Training procedure.}\quad The two learning-modules used in our approach, namely \emph{Dense Descriptor Extraction} and \emph{Depth Estimation}, are both self-supervised in that they can be trained on video sequences with corresponding SfM results obtained using a conventional hand-crafted feature descriptor. This training strategy is introduced in~\cite{liu2019dense,liu2020extremely}. Before training the complete pipeline here, we assume that both the aforementioned modules were pre-trained using such a self-supervised strategy. Then, the training order is as follows. First, the pre-trained dense descriptor extraction network is first used to establish correspondences that produce an SfM result. If the result is unsatisfactory, the dense descriptor extraction network will be fine-tuned with this SfM result for bootstrapping. This process can be repeated if necessary. We found the pre-trained dense descriptor extraction network to generalize well to unseen videos so that iterative fine-tuning was not required. Then, the depth estimation network is fine-tuned using the patient-specific dense descriptor extractor and SfM results to achieve the best performance on the input video. Each module in the pipeline is introduced below, please refer to the supplementary material for details of the implementation such as network architecture and loss design.

\begin{figure*}[t]
\centering
\includegraphics[width=0.85\textwidth]{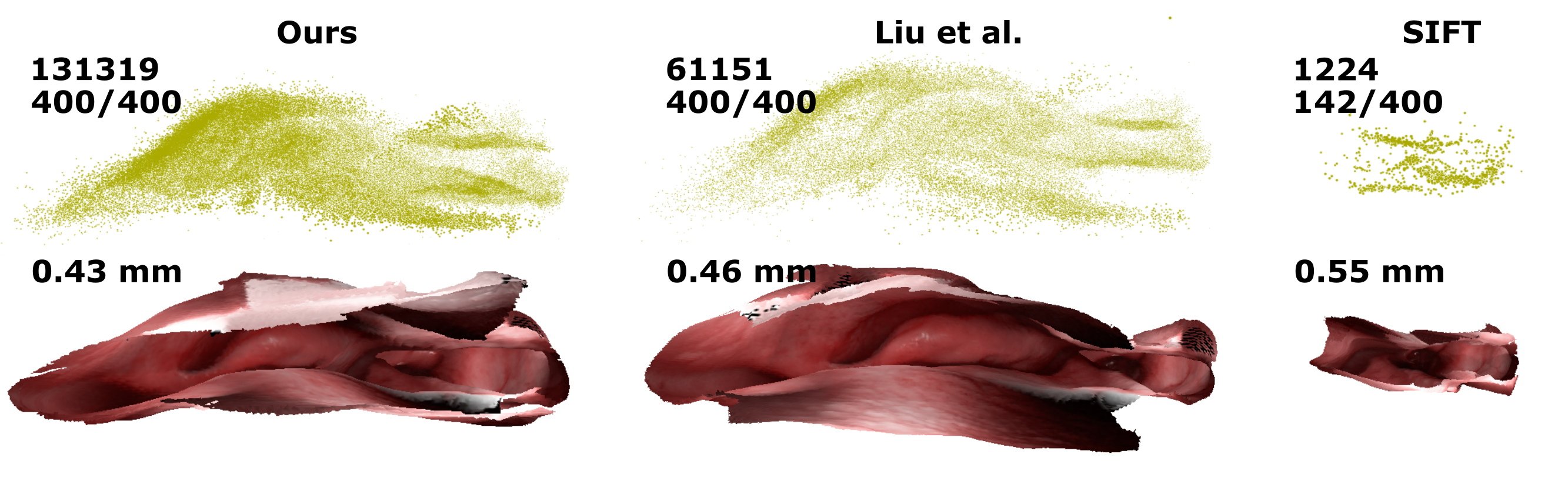}
\caption{\textbf{Comparison of reconstruction with different descriptors.}\quad We compared the sparse and surface reconstruction using our proposed dense descriptor with those using the descriptor from Liu~\etal~\cite{liu2020extremely} and SIFT~\cite{lowe2004distinctive}. The first row shows sparse reconstructions from SfM using different descriptors. The second row displays surface reconstructions estimated using the proposed method based on the sparse reconstruction. For each column, from top to bottom, the three numbers correspond to the number of points in the sparse reconstruction, number of registered views out of the total ones, and the point-to-mesh distance between the sparse and surface reconstruction. The first row shows that our sparse reconstruction is two times denser than~\cite{liu2020extremely}. Surface reconstruction with SIFT covers much less area and has high point-to-mesh distance, which shows the importance of having a dense enough point cloud and complete camera trajectory.}
\label{fig:descriptor_comparison}
\end{figure*}

\textbf{Structure from Motion with Dense Descriptor.}\quad SfM simultaneously estimates a camera trajectory and a sparse reconstruction from a video. We choose SfM over other multi-view reconstruction methods such as SLAM because SfM is known to produce more accurate reconstruction at the cost of increased run time. Still, it has been shown in~\cite{liu2020extremely} that local feature descriptors have difficulty in dealing with smooth and repetitive textures that commonly occur in endoscopy. In this work, we adopt the learning-based dense descriptor extraction method in~\cite{liu2020extremely} to replace the role of local descriptors for pair-wise feature matching in SfM. Intuitively, such a learning-based approach largely improves the matching performance because the Convolutional Neural Network-based architecture enables global context encoding. In addition, the pixel-wise feature descriptor map generated from the network also enables dense feature matching, which eliminates the reliance on repeatable keypoint detections. With a large number of correct pair-wise point correspondences being found, the density of the sparse reconstruction and the completeness of the camera trajectory estimate are largely improved compared with SfM with local descriptors. For each query location in the source image that is suggested by a keypoint detector, a matching location in the target image is searched for. By comparing the feature descriptor of the query location to the pixel-wise feature descriptor map of the target image, a response map is generated. In order to achieve subpixel matching accuracy, we further apply bicubic interpolation to the response map and use the position with the maximum response as the matching location. A comparison of descriptors and the impact on surface reconstruction is shown in Fig.~\ref{fig:descriptor_comparison}.

\textbf{Depth Estimation.}\quad Liu~\etal~\cite{liu2019dense} proposed a method that can train a depth estimation network in a self-supervised manner with sparse guidance from SfM results and dense inter-frame geometric consistency. In this work, we adopt a similar self-supervision scheme and the network architecture as~\cite{liu2019dense} but assume that depth estimates should be probabilistic because poorly illuminated areas will likely not allow for precise depth estimates. Consequently, we model depth as a pixel-wise independent Gaussian distribution that is represented by a mean depth and its standard deviation. The related training objective is to maximize the joint probability of the training data from SfM given the predicted depth distribution. The probabilistic strategy provides some robustness to outliers from SfM, as shown in Fig.~\ref{fig:depth_comparison}. We also add an appearance consistency loss~\cite{zhou2017unsupervised}, which is commonly used in self-supervised depth estimation for natural scenes where photometric constancy assumptions are reasonable. This assumption, however, is invalid in endoscopy and cannot be used for additional self-supervision. Interestingly, the pixel-wise descriptor map from \emph{Dense Descriptor Extraction} module is naturally illumination-invariant and provides a dense signal. It can thus be interpreted in analogy to appearance consistency, where appearance is now defined in terms of descriptors rather than raw intensity values. This seemed to further improve the performance, which is qualitatively shown in Fig.~\ref{fig:depth_comparison}. Based on the sparse supervision from SfM together with the dense constraints of geometric and appearance consistency, the network learns to predict accurate dense depth maps with uncertainty estimates for all frames, which are fused to form a surface reconstruction in the next step.

\textbf{Depth Fusion and Surface Extraction.}\quad We apply a depth fusion method \cite{curless1996volumetric} based on truncated signed distance functions~\cite{zach2007globally} to build a volumetric representation of the sinus surface. Depth measurements are propagated to a 3D volume using ray-casting from the corresponding camera pose and the corresponding uncertainty estimates determine the slope of the truncated signed distance function for each ray. We used SfM results to re-scale all depth estimates before the fusion to make sure all estimates are scale-consistent. To fuse all information correctly, the camera poses estimated from SfM are used to propagate the corresponding depth estimates and color information to the 3D volume. Finally, the Marching Cubes method~\cite{lorensen1987marching} is used to extract a watertight triangle mesh surface from the 3D volume.

\section{Experiments}
\textbf{Experiment Setup.}\quad The endoscopic videos used in the experiments were acquired from eight consenting patients and five cadavers under an IRB approved protocol. The anatomy captured in the videos is the nasal cavity. The total time duration of videos is around $40$ minutes. Because this method is patient-specific, all data are used for training. All processing related to the proposed pipeline used $4$-time spatially downsampled videos, which have a resolution of $256\times 320$. SfM was first applied with SIFT~\cite{lowe2004distinctive} to all videos to generate sparse reconstructions and camera trajectories. Results of this initial SfM run were used to pre-train the depth estimation and dense descriptor extraction networks until convergence. Note that the pre-trained depth estimation network was not trained with appearance consistency loss. For evaluation of each individual video sequence, SfM was applied again with the pre-trained dense descriptor extraction network to generate a denser point cloud and a more complete camera trajectory. The depth estimation network, now with appearance consistency loss, were fine-tuned with the updated and sequence-specific SfM results. Note that if the pre-trained descriptor network cannot produce satisfactory SfM results on the new sequence, descriptor network fine-tuning and an extra SfM run with fine-tuned descriptor are required. All experiments were conducted on one NVIDIA TITAN X GPU. The registration algorithm used for evaluation is based on~\cite{billings2015generalized} to optimize over similarity transformation.

\begin{figure*}[t]
\centering
\includegraphics[width=0.75\textwidth]{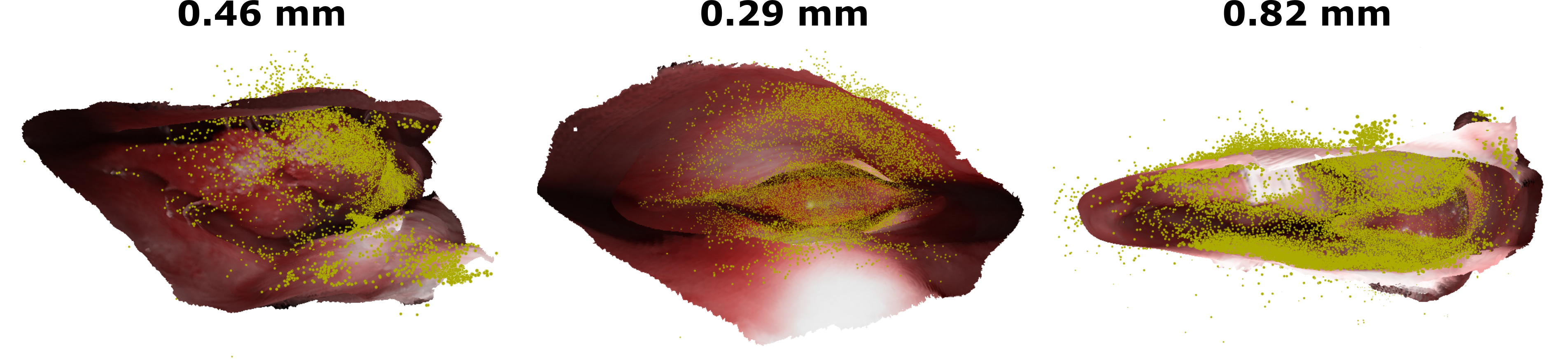}
\caption{\textbf{Overlay of sparse and surface reconstruction.}\quad Sparse reconstruction from SfM is overlaid with surface reconstruction from the pipeline. The number in each column represents the average point-to-mesh distance.}
\label{fig:point_cloud_overlay}
\end{figure*}

\textbf{Agreement with SfM results.}\quad Because our method is self-supervised and SfM results are used to derive supervisory signals, the discrepancy between the surface and sparse SfM reconstruction should be minimal. To evaluate the consistency between our surface reconstruction and the sparse reconstruction, we calculated the point-to-mesh distance between the two. Because scale ambiguity is intrinsic for monocular-based surface reconstruction methods, we used the CT surface models to recover the actual scale for all individuals where CT data are available. For those that do not have corresponding CT data, we used the average statistics of the population to recover the scale. The evaluation was conducted on $33$ videos of $13$ individuals. The estimated point-to-mesh distance was $0.34\left(\pm0.14\right)\,\mathrm{mm}$. Examples of the sparse and surface reconstruction overlaid with point-to-mesh distance are shown in Fig.~\ref{fig:point_cloud_overlay}.

\textbf{Consistency against video variation.}\quad Surface reconstruction methods should be insensitive to variations in video capture, such as camera speed. To evaluate the sensitivity of our method, we randomly sub-sampled frames from the original video to mimic camera speed variation. The pipeline was run for each sub-sampled video and we evaluated the model consistency by aligning surface reconstructions estimated from different subsets. To simulate camera speed variation, out of every $10$ consecutive video frames, only $7$ frames were randomly selected. We evaluated the model consistency on $3$ video sequences that cover the entire nasal cavity of three individuals, respectively. Five reconstructions that were computed from random subsets of each video were used for evaluation. The average residual distance after registration between different surface reconstructions was used as the metric for consistency. The scale recovery method is the same as above. The residual error was $0.21\left(\pm0.10\right)\,\mathrm{mm}$.

\begin{figure*}[t]
\centering
	\includegraphics[width=0.8\textwidth]{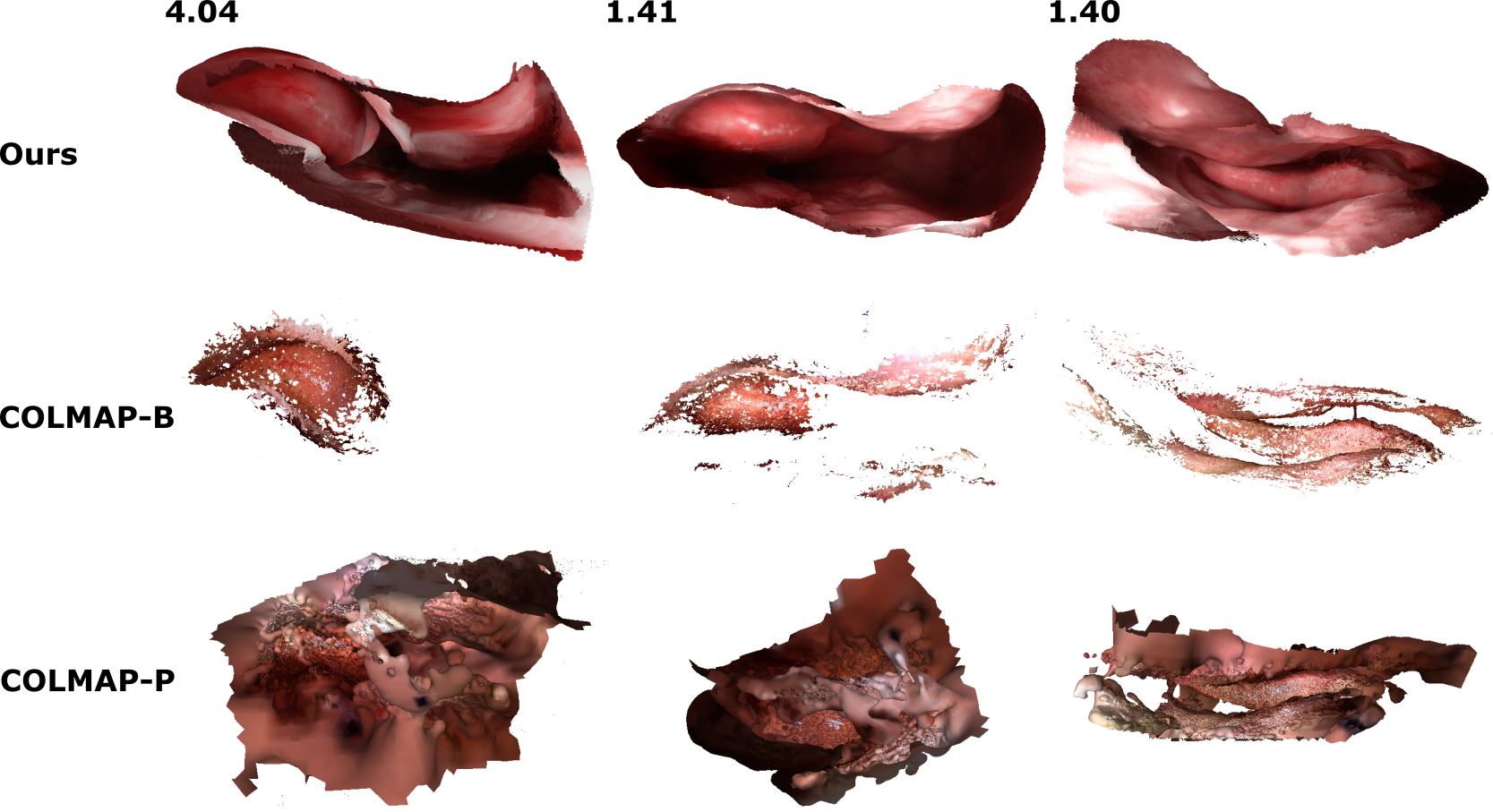}
	\caption{\textbf{Comparison of surface reconstruction from ours and COLMAP.}\quad The number in each column is the ratio of surface area between our reconstruction and COLMAP with ball pivoting~\cite{Bernardini1999Ball} (COLMAP-B). Ratios are underestimated because many redundant invalid surfaces are generated in the second row. COLMAP with Poisson~\cite{kazhdan2006poisson} (COLMAP-P) is shown in the last row with excessive surfaces removed already. 
	}
	\label{fig:reconstruction_method_comparison}
\end{figure*}

\textbf{Agreement with dense reconstruction from COLMAP.}\quad We used the ball pivoting~\cite{Bernardini1999Ball} method to reconstruct surfaces in COLMAP instead of built-in Poisson~\cite{kazhdan2006poisson} and Delaunay~\cite{cazals2006delaunay} methods because these two did not produce reasonable results. Three videos from $3$ individuals were used in this evaluation. The qualitative comparison is shown in Fig.~\ref{fig:reconstruction_method_comparison}. The same scale recovery method as above was used. The average residual distance after registration between the surface reconstructions from the proposed pipeline and COLMAP is $0.24\left(\pm0.08\right) \mathrm{mm}$. In terms of the runtime performance, given that a pre-trained generalizable descriptor network and depth estimation network exist, our method requires running sparse SfM with a learning-based feature descriptor, fine-tuning depth estimation network, depth fusion, and surface extraction. For the three sequences, the average runtime for the proposed method is $127$ minutes, whereas the runtime for COLMAP is $778$ minutes.

\begin{figure}[t]
\begin{subfigure}{0.622\textwidth}
  \centering
  \includegraphics[width=\textwidth]{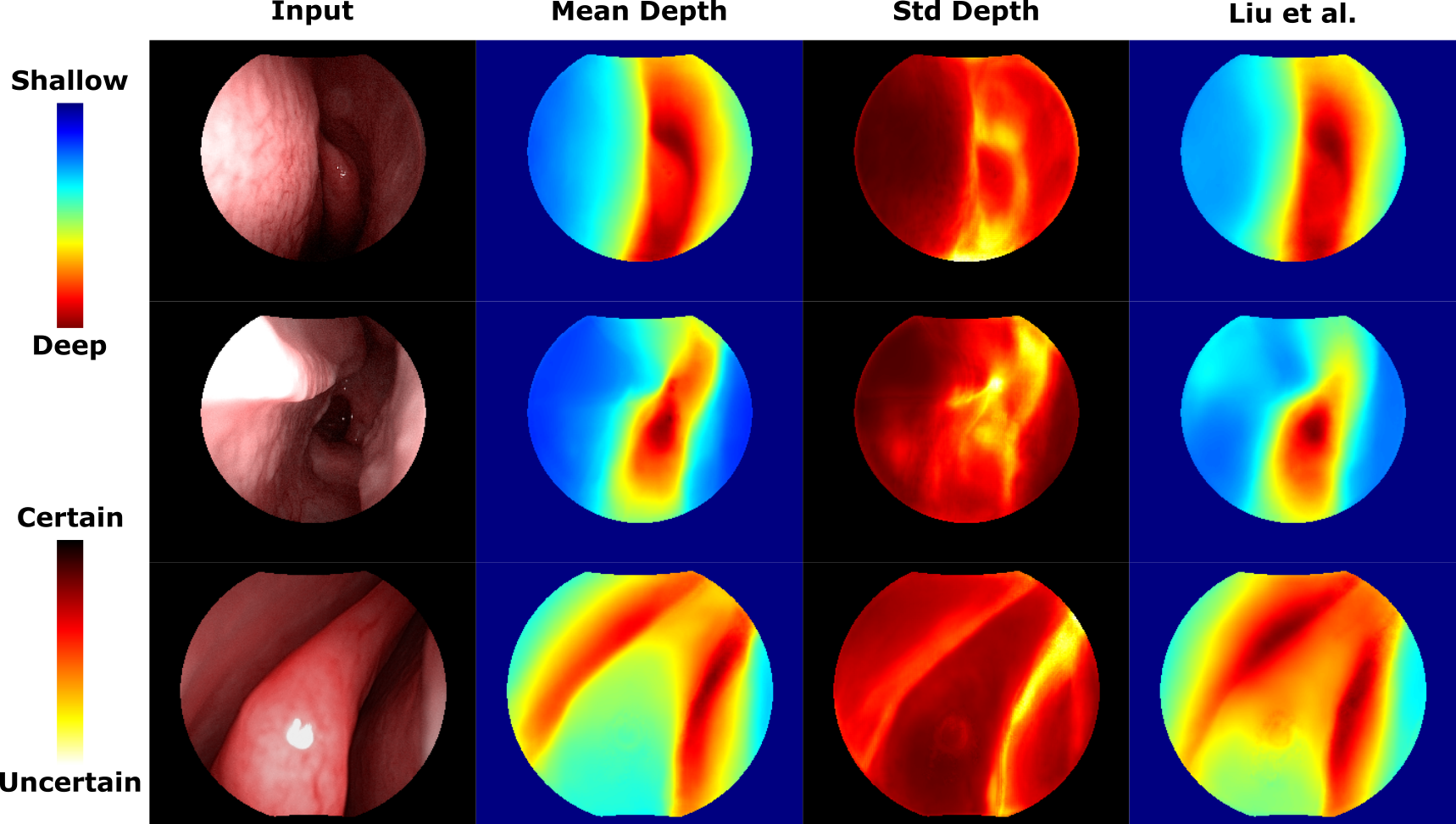}  
  \caption{}
  \label{fig:depth_comparison}
\end{subfigure}
\begin{subfigure}{0.37\textwidth}
  \centering
  \includegraphics[width=\textwidth]{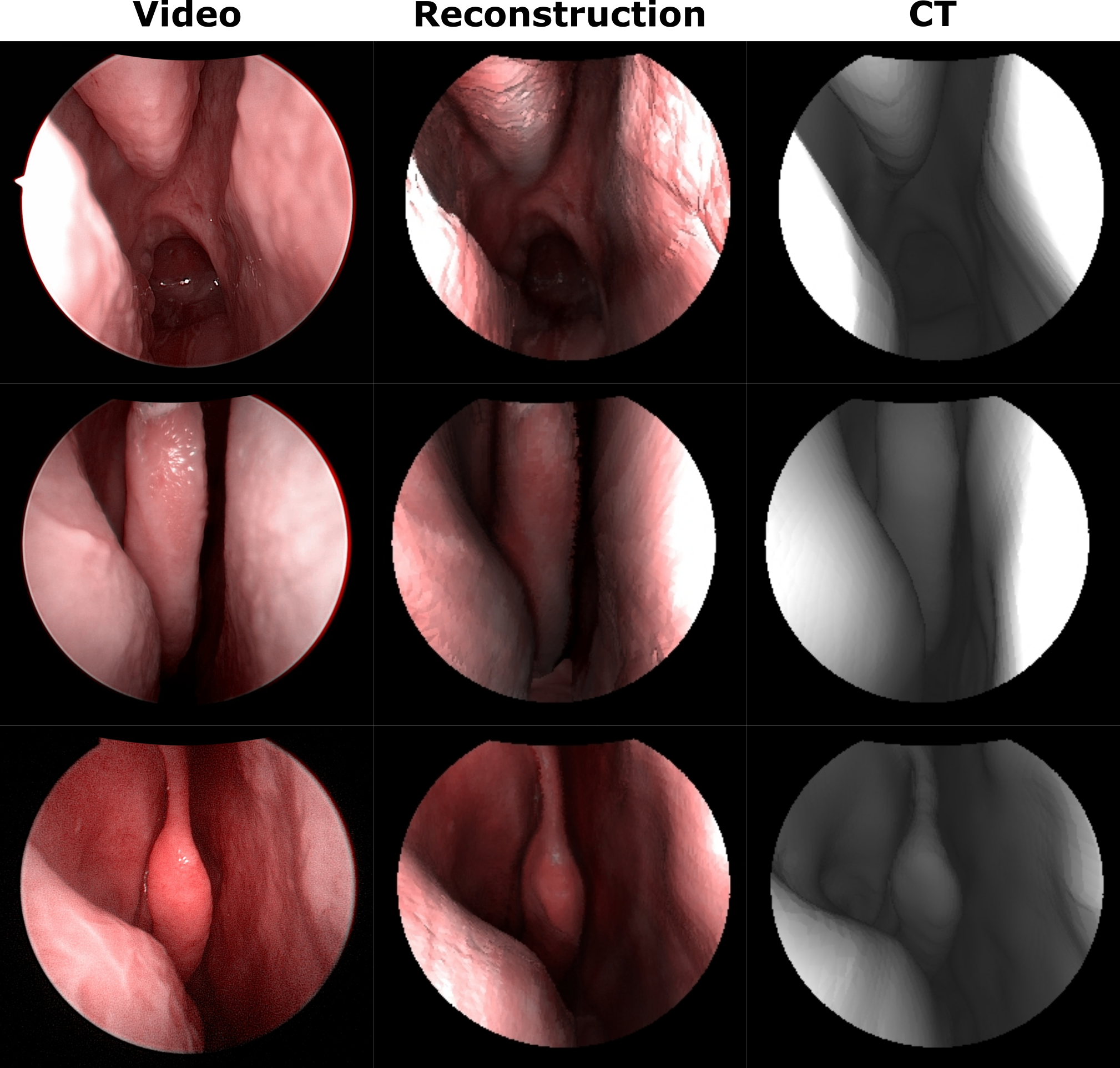} 
  \caption{}
  \label{fig:CT_recon_alignment}
\end{subfigure}
\caption{\textbf{(a) Comparison of depth estimation.}\quad By enforcing inter-frame appearance consistency during network training and introducing depth uncertainty estimation, the depth predictions seem to be visually better compared to those from the model trained with the settings in~\cite{liu2019dense}. As can be seen in the third column, higher depth uncertainties were predicted for deeper regions and those where training data is erroneous, such as the specularity in the last row. \textbf{(b) Visualization of aligned endoscopic video, dense reconstruction, and CT surface model.}\quad To produce such visualization, our dense reconstruction is first registered to the CT surface to obtain the transformation between the two coordinate systems.}
\end{figure}

\textbf{Agreement with CT.}\quad Model accuracy was evaluated by comparing surface reconstructions with the corresponding CT models. In this evaluation, two metrics were used: average residual error between the registered surface reconstruction and the CT model, and the average relative difference between the corresponding cross-sectional areas of the CT surface models and the surface reconstructions. The purpose of this evaluation is to determine whether our reconstruction can be used as a low-cost, radiation-free replacement for CT when calculating clinically relevant parameters. To find the corresponding cross-section of two models, the surface reconstruction was first registered to the CT model. The registered camera poses from SfM were then used as the origins and orientations of the cross-sectional planes. The relative differences of all cross-sectional areas along the registered camera trajectory were averaged to obtain the final statistics. This evaluation was conducted on $7$ video sequences from $4$ individuals. The residual error after registration was $0.69\left(\pm0.14\right)\,\mathrm{mm}$. As a comparison, when the sparse reconstructions from SfM are directly registered to CT models, the residual error was $0.53\left(\pm0.24\right) \mathrm{mm}$. The smaller error is due to the sparsity and smaller region coverage of the sparse reconstruction compared to ours. In Fig.~\ref{fig:CT_recon_alignment}, a visualization of the video-reconstruction-CT alignment is shown. The cross-sectional surface areas are estimated with an average relative error of $7\left(\pm2\right)$\,\%. This error mainly originates from regions that were not sufficiently visualized during scoping, such as the inferior, middle, and superior meatus. These regions are included in our analysis due to the automation of cross-sectional measurements. In practice, these regions are not commonly inspected as they are hidden beneath the turbinates; if a precise measurement of these areas is desired, small modifications to video capture would allow for improved visualization. Similar to~\cite{ma2019real}, such adjustments can be guided by our surface reconstruction, since the occupancy states in the fusion volume can indicate explicitly what regions were not yet captured with endoscopic video.

\section{Discussion}
\textbf{Choice of depth estimation method.}\quad In this work, a monocular depth estimation network is used to learn the complex mapping between the color appearance of a video frame and the corresponding dense depth map. The method in~\cite{liu2019dense} has been shown to generalize well to unseen cases. However, the patient-specific training in this pipeline may allow for higher variance mappings since it does not need to generalize to other unseen cases. Therefore, a more complex network architecture could potentially further improve the depth estimation accuracy, leading to more accurate surface reconstruction. For example, a self-supervised recurrent neural network that predicts the dense depth map of a video frame based on the current observation and the previous frames in the video could potentially have more expressivity and be able to learn a more complex mapping, such as the method proposed by Wang~\etal~\cite{wang2019recurrent}.

\textbf{Limitations.}\quad First, this pipeline will not work if SfM fails. This could happen in some cases, such as in the presence of fast camera movement, blurry images, or tissue deformation. The latter may potentially be tackled by non-rigid SfM~\cite{Khan2018Robust}. Second, the pipeline currently does not estimate geometric uncertainty of the surface reconstruction. A volume-based surface uncertainty estimation method may need to be developed for this purpose.

\section{Conclusion}
In this work, we proposed a learning-based surface reconstruction pipeline for endoscopy. Our method operates directly on raw endoscopic videos and produces watertight textured surface models that are in good agreement with anatomy extracted from CT. While this method so far has only been evaluated on videos of the nasal cavity, the proposed modules are generic, self-supervised, and should thus be applicable to other anatomies. Future work includes uncertainty estimation on the reconstructed surface models and prospective acquisition of longitudinal endoscopic video data in the clinic.

\bibliographystyle{splncs}
\bibliography{main}

\end{document}